\documentclass[letterpaper, 10 pt, conference]{ieeeconf}

\IEEEoverridecommandlockouts  
\makeatletter
\let\NAT@parse\undefined
\makeatother

\usepackage{graphicx}
\usepackage{booktabs}
\usepackage{hyperref}
\usepackage{amsmath}
\usepackage[dvipsnames]{xcolor}
\usepackage[numbers]{natbib}
\usepackage[scaled]{beramono}
\usepackage[T1]{fontenc}

\usepackage{tabularx}
\usepackage{multirow}

\newcommand{\code}[1]{{\texttt{#1}}}

\title{\LARGE\bf {\sc{CAPE}}: Corrective Actions from Precondition Errors \\ using Large Language Models}

\author{
  Shreyas Sundara Raman$^{1*}$, Vanya Cohen$^{2}$, Ifrah Idrees$^{1}$, Eric Rosen$^{1}$, \\ Ray Mooney$^{2}$, Stefanie Tellex$^{1}$, David Paulius$^{1}$
  \thanks{Project Website: \url{https://shreyas-s-raman.github.io/CAPE/}}
  \thanks{$^{*}$Corresponding Author (Email: \href{mailto:shreyas_sundara_raman@brown.edu}{\texttt{shreyas\textunderscore sundara\textunderscore raman@brown.edu}})}
  \thanks{$^{1}$Brown University, Providence, RI, USA.}
  \thanks{$^{2}$The University of Texas at Austin, Austin, TX, USA.}
}

\begin{document}

\maketitle
\thispagestyle{empty}
\pagestyle{empty}

\thispagestyle{plain}
\pagestyle{plain}


\begin{abstract}

Extracting commonsense knowledge from a large language model (LLM) offers a path to designing intelligent robots. Existing approaches that leverage LLMs for planning are unable to recover when an action fails and often resort to retrying  failed actions, without resolving the error's underlying cause. 
We propose a novel approach (CAPE) that attempts to propose corrective actions to resolve precondition errors during planning. CAPE improves the quality of generated plans by leveraging few-shot reasoning from action preconditions. 
Our approach enables embodied agents to execute more tasks than baseline methods while ensuring semantic correctness and minimizing re-prompting. In VirtualHome, CAPE generates executable plans while improving a human-annotated plan correctness metric from 28.89\% to 49.63\% over SayCan. Our improvements transfer to a Boston Dynamics Spot robot initialized with a set of skills (specified in language) and associated preconditions, where CAPE improves the correctness metric of the executed task plans by 76.49\% compared to SayCan. Our approach enables the robot to follow natural language commands and robustly recover from failures, which baseline approaches largely cannot resolve or address inefficiently.
%
%
\end{abstract}


\section{Introduction}
\label{introduction}

\begin{figure*}[ht]
    \centering
    \includegraphics[width=\textwidth]{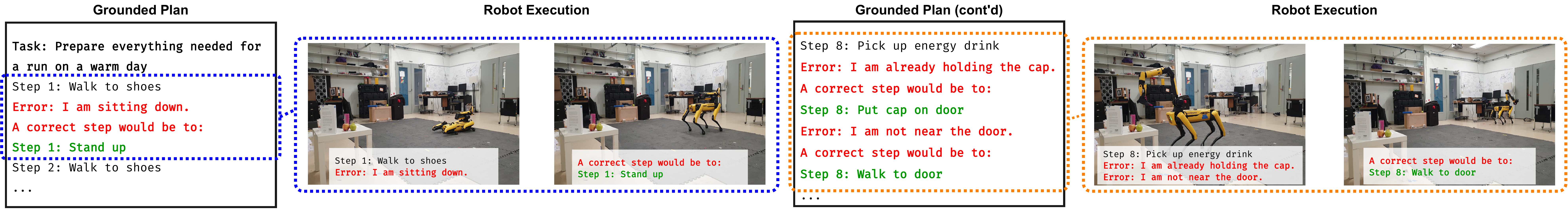}
    \caption{Qualitative results of CAPE for robot execution of the task \textit{"prepare for a run"}. We highlight $2$ cases where re-prompting with precondition error information resolves action failures (\textit{left:} resolving prerequisite for walking by standing; \textit{right:} resolving one-armed manipulation constraint).}
    \vspace{-0.35cm}
    \label{fig:robot-demo}
\end{figure*}

Generalized robots can assist humans by accomplishing a diverse set of goals in varying environments. Many such agents are equipped with a library of skills for primitive action execution. Here, natural language can enable more seamless human-robot interaction by leveraging these skill libraries~\cite{tellex2020robots}. Given a task description or command from a human, a robot must be able to autonomously propose a sequence of actions (from its skill repertoire) that realizes the given task. Critical to such an application is the agent's ability to ground skills specified in language to their environment and reason about state changes from skill execution or the relevance of proposed actions towards a task's objective. For instance, if a robot is commanded to ``put away groceries'', it must ground the concept of ``groceries'' to objects in its environment and decompose the task of ``putting away'' to meaningful constituent skills from its repertoire. 

Thus, extracting actionable knowledge from LLMs requires context about the agent's embodiment and environment state. 
Related works that extract plans from LLMs using prompting strategies assume access to extra information such as: 1) predefined skills with preconditions~\cite{saycan_corl} 2)visual-language models that determine affordance from observations like SayCan~\cite{saycan_corl}, 3) descriptions of the agent's goal~\cite{huang2022language,huang2022inner} or 4) descriptions of observation and action spaces for reasoning in text-based video games~\cite{yao2020calm,singh2021pretrained}. 
These approaches do not efficiently nor explicitly resolve failure modes during planning: they either propose actions that are not afforded execution in the environment (i.e. violate preconditions, such as walking through a closed door), or resort to exploring the entirety of an agent's action library to identify affordable actions~\cite{saycan_corl}.

%

\begin{figure*}[t]
    \centering
    \includegraphics[width=\textwidth]{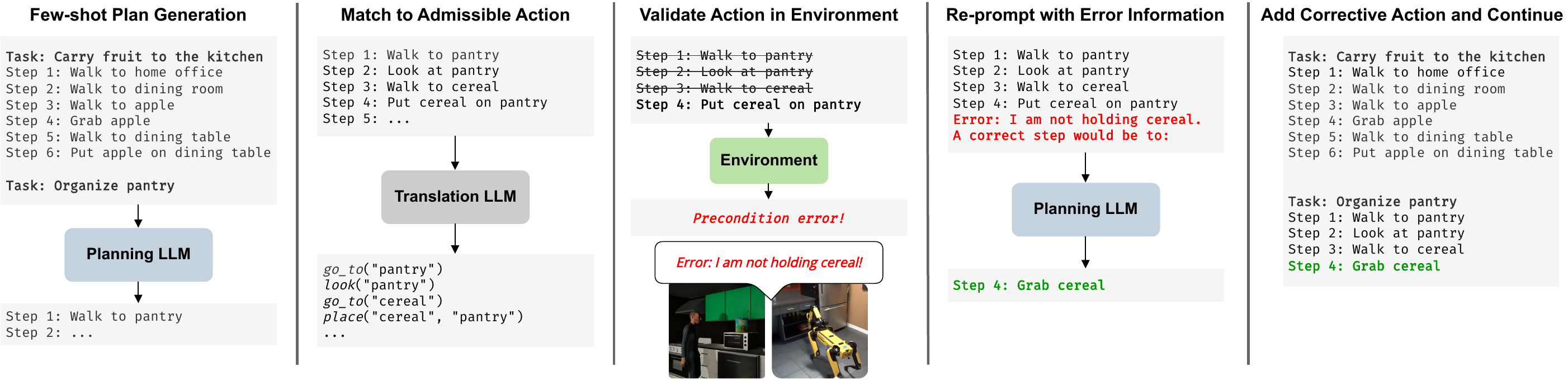}
    \vspace{-0.5cm}
    \caption{Overview of CAPE: To generate an executable plan, we select an in-context example task (from a demonstration set) that is most semantically similar to the query task. The Planning LLM generates a natural language description for the next step in the plan. The Translation LLM~\cite{liu2019roberta} grounds this description to an admissible skill in the agent's repertoire. If this action violates preconditions for the proposed skill, the precondition error information is formatted into a \textit{corrective prompt}, which along with the failed skill are provided to the LLM for corrective action proposal.}    
    \vspace{-0.35cm}
    \label{fig:our-pipeline}
\end{figure*}

We use \textit{precondition errors} to resolve action failure, which is motivated by the vast body of research on planning algorithms and definitions like PDDL~\cite{mcdermott1998pddl}. In these settings, robots are equipped with a repertoire of skills, each requiring certain \textit{preconditions} to be satisfied in order to afford their execution. We target the failure mode of executing skills without satisfying their preconditions in this setting.
Using parametrized skills that are codified into natural language, we leverage a LLM to generate a sequence of actions for execution towards completing a task. When a robot or agent fails to execute an action due to precondition violation, we use a templated-prompting strategy called CAPE (\textit{Corrective Actions from Precondition Errors}) to query the LLM for corrective actions (Figure~\ref{fig:our-pipeline}). Our prompts either specify that the action failed or provide explanatory details about the cause of action failure, flexible to the extent of knowledge accessible to the robot about its skills or domain. This paper builds on our previous work \cite{raman2022planning} with more rigorous analysis, larger scale human evaluation, additional (more competitive) baselines and experiments both in simulation and real-world settings.

Our contributions are as follows: we introduce CAPE a novel approach for LLM planning that generates corrective actions to recover from failure, using prompts based on precondition errors and few-shot learning. We detail how our re-prompting strategy can be deployed on embodied systems with both large and small skill repertoires using different re-prompting methods. We also evaluate CAPE against several baselines~\cite{huang2022language,saycan_corl} and ablations to show our method achieves near-perfect plan executability and more semantically correct plans for various tasks executed on a Boston Dynamics Spot robot and a simulated agent in VirtualHome~\cite{puig2018virtualhome}.


\vspace{-0.1cm}
\section{Background}
\label{background}

\textbf{In-Context Learning:}
%
%
\citet{brown2020language} introduced GPT-3: a 175 billion parameter language model capable of few-shot learning for novel tasks, including Q\&A, arithmetic, and comprehension, by prompting the LLM with in-context task examples used for structural and syntactic guidance.
This approach offers several advantages over task learning with fine-tuned pre-trained latent language representations~\cite{radford2018improving,devlin2018bert,peters2018deep} and zero-shot inference~\cite{radford2019language} due to sample efficiency and task generalization.
In-context learning performs best when examples are relevant to the test task; we retrieve in-context examples based on their semantic similarity to a task~\cite{liu2021makes,huang2022language}.

\textbf{Open-Loop Plan Generation:}
\label{open-loop}
CAPE extends the open-loop framework of \citet{huang2022language}, which generates a plan for a task zero-shot without feedback from the environment.
Given a query task $\mathcal{Q}$ (i.e. the target task), first, a high-level example task $\mathcal{T}$ and its plan are chosen from a \textit{demonstration set} as a contextual example of a free-form plan for the \emph{Planning LLM}; note that $\mathcal{T}$ is selected to maximizes cosine similarity with the \textit{query task} $\mathcal{Q}$. The \emph{Planning LLM} auto-regressively generates actions for task $\mathcal{Q}$ in free-form language via in-context learning.
The \emph{Translation LLM} then utilizes a BERT-style LM (Sentence-BERT~\cite{liu2019roberta}) to embed the generated free-form actions ($a_l$) to the most semantically (i.e., cosine-similar) action in the agent's repertoire ($a_e$). Here, an admissible action refers to a language description of an action in the agent's skill repertoire. 
The chosen admissible action ($\hat{a}_e$) is then appended to the unfinished prompt to condition future auto-regressive step generation on admissible actions. We investigate how to improve planning in the closed-loop domain by leveraging precondition error feedback as an auxiliary modality of information.

\textbf{Affordance and Preconditions:}
Action preconditions and effects are commonly adopted in robot planning domains, such as those using PDDL~\cite{mcdermott1998pddl} or STRIPS~\cite{fikes1971strips}, where a set of predefined skills are accessible to robots.
Structured affordance models factorize states into \textit{preconditions}, which define affordance by independent state components that must be satisfied for execution. 
This can be formalized by the options framework~\cite{sutton1999between}, where options $\mathcal{O}(s)$ over the state space $\mathcal{S}$ form a set of temporally extended actions equivalent to those in an agent's skill repertoire. An initiation set of an option $\mathcal{I}(o)$ defines the states in which option execution is afforded (akin to preconditions), while a termination condition $\beta_{o}(s)$ describes the terminal state of the skill. If the current state fails to meet the initiation state of an option, a precondition error arises.
%
Environment states in these domains can be factorized in a semantically meaningful manner to evaluate the validity of preconditions for a skill, thus enabling a skill's affordance to be measured.
Learning and modeling preconditions have been largely studied in model-based approaches that leverage symbolic planning~\cite{konidaris2018skills,garrett2021integrated}. Our work investigates how these preconditions can be leveraged to improve planning using LLMs.

\section{Method}
\label{method}


Given a task specified in natural language, we use LLMs to generate a plan. When an agent or robot fails skill execution, CAPE integrates precondition errors into a prompt that aims to repair plans.

\subsection{Plan Generation via Re-prompting}
\label{sec:reprompt}

\begin{figure}[t]
    \centering
    \includegraphics[width=0.6\columnwidth]{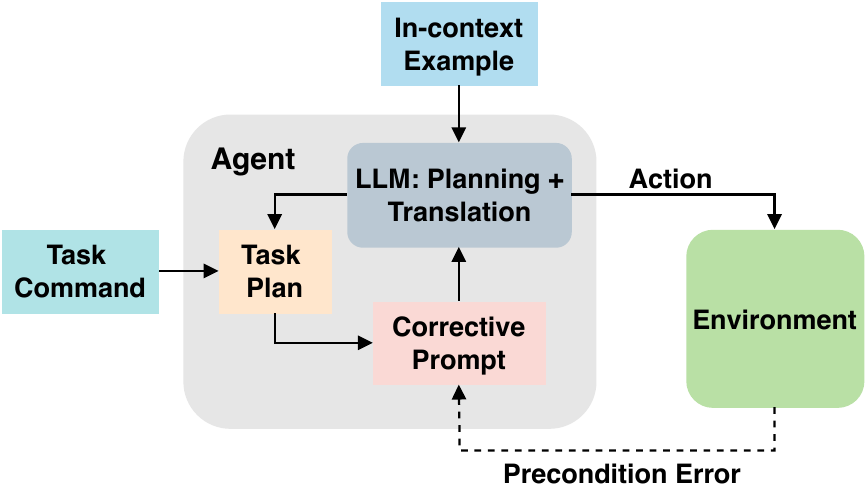}
    \caption{CAPE uses a LLM to generate plans for tasks specified in natural language. When the agent fails to execute a step, we re-prompt the LLM with error information, utilizing the latent commonsense reasoning and few-shot learning capabilities of the LLMs to overcome execution errors.}
    \vspace{-0.5cm}
    \label{fig:mini-pipeline}
\end{figure}

In control theory, a closed-loop system relies on feedback from its outputs for adaptive control~\cite{golnaraghi2017automatic}. Similarly, CAPE leverages error feedback in a closed-loop planning setup, which allows it to correct a generated plan when any action proposed by the LLM is not afforded execution, by injecting precondition error information as \textit{corrective prompts} (see Figure~\ref{fig:mini-pipeline}).
Certain errors require more context about the agent's state, action history and environment. For instance, correcting an error in VirtualHome~\cite{puig2018virtualhome} such as \code{<character> (1) does not have a free hand when executing "[GRAB] <obj> (1) [1]"} requires knowledge of what objects the agent previously grabbed or is currently holding, as well as available adjacent objects on which to drop the held object and free the agent's hands. We construct corrective prompts composed of the following segments of information:
\begin{itemize}
    \item \textbf{Contextual Information}: This includes relevant context and action history upon action failure. We supply the query task $\mathcal{Q}$ and the query steps up to the action that has failed for context.
    \item \textbf{Precondition Error Information}: We optionally include details on the violated precondition in the prompt, which is tailored based on the degree to which the agent can assess precondition violations.
\end{itemize}


In order for the Translation LLM to ground the natural utterance, we need to assume that the agent is equipped with a skill repertoire of actions that are admissible to the environment. Thus, preconditions only need to be defined for each general parametrized skill. It is important to note that the Planning LLM used by CAPE does not explicitly know about the agent's skills nor the preconditions for each skill during the re-prompting process. Instead, we utilize the preconditions (a set of logical propositions assessing a skill's affordance) defined for each parametrized skill in our skill repertoire to obtain precondition errors by comparing with the environment's current state. The environment state and precondition propositions are external to the LLM, but the error information produced by them can then be integrated into a corrective language prompt. As a result, there is a significant layer of abstraction, where the Planning LLM has to \textit{infer} the cause of failures and environment mechanics based only on the context provided by the corrective prompt and the agent's own action history in order to propose an appropriate corrective action. The use of preconditions is typical in planning domains where the robot or agent has skills built on representations that define preconditions and effects, e.g., PDDL~\cite{mcdermott1998pddl}, STRIPS~\cite{fikes1971strips} or LTL~\cite{pnueli1977temporal}. Since preconditions are already defined in these representations, appropriate language feedback can be integrated into the precondition module with minimal extra effort.


\subsubsection*{Re-prompting Strategies}
We re-prompt with varying degrees of precondition error detail in both zero-shot ($\mathcal{Z}$) and few-shot ($\mathcal{F}$) approaches, and denote either setting by ${P}, \text{where } P = \mathcal{Z}\lor\mathcal{F}$. 
Few-shot reprompting ($\mathcal{F}$) provides $3$ incontext precondition errors and corrective actions, taken from the demonstration set that is separate from the query task, that are only injected when the LLM-Agent needs to propose corrective actions i.e. not for executable actions. Re-prompting strategies can be categorized as follows:


\begin{itemize}
    \item {\textbf{Re-prompting with Success Only ($\mathcal{Z}_{S}$):} solely informs the LLM that the action failed (i.e., \code{``Task Failed''}).\footnote{This is analogous to success detection used in Inner Monologue~\cite{huang2022inner}, which was used to determine whether to re-execute failed actions since low-level policy success is stochastic. However, our aim is to repair the high-level plans generated by the LLM with corrective actions that arise from a new distribution of actions using precondition feedback.}}
    \item {\textbf{Re-prompting with Implicit Cause ($\mathcal{Z}_{I}$):} provides more detail to the LLM with a prompt template containing the name of the failed action and the object(s) the agent interacted with (i.e., \code{``I cannot <action> <object>''}). This requires the LLM to infer the cause of error when proposing corrective actions.}
    \item {\textbf{Re-prompting with Explicit Cause ($\mathcal{Z}_{E}$):}  states the precondition violation that prevents action execution, in addition to feedback provided by $\mathcal{Z}_{I}$ (i.e., \code{``I cannot <action> <object> because <precondition violation>''}).}
\end{itemize}



$P_{E}$ gives the most error feedback to the LLM. However, $P_{S}$ and $P_{I}$ only require a target object and skill associated with the failed action, which the LLM proposes.
Likewise, a $\mathcal{P}_{S}$ prompt can work with visual-language model approaches like SayCan~\cite{saycan_corl}, whereas $P_{I}$ and $P_{E}$  can work with task and motion planning approaches~\cite{garrett2021integrated}).

\subsubsection*{Scoring Grounded Actions}
We use the scoring function $\mathcal{S}_w$ (Equation~\ref{eq:original_score}), a weighted combination of log probability and cosine similarity, which is thresholded to determine the feasibility of each proposed grounded step~\cite{huang2022language}. 
Log probability is defined as $P_\theta(X_i) := \frac{1}{n_i}\sum_{j=1}^{n_i}$log$p_\theta(x_{i,j}|x_{i<j})$, where $\theta$ parameterizes the pretrained Planning LLM and $X_i$ is a generated step consisting of $n$ tokens $(x_{i,1},...,x_{i,n})$. 
Cosine similarity is defined as $C(f(\hat{a}),f(a_e)):=\frac{f(\hat{a})\cdot f(a_{e})}{||f(\hat{a})||||f(a_{e})||}$, where $f$ is the Translation LLM embedding function, $a$ is the predicted action, and $a_e$ is the admissible action for which we estimate the distance with respect to:
\begin{equation}
    \label{eq:original_score}
    \mathcal{S}_{w} = \underset{a_e}{\operatorname{argmax}}~[\underset{\hat{a}}{\operatorname{max}} ~C(f(\hat{a}),f(a_{e})) + \beta\cdot P_{\theta}(\hat{a})],
\end{equation}
where $\beta$ is a weighting coefficient.
$\mathcal{S}_{w}$ prioritizes the quality of natural language at the cost of semantic translation and often results in mistranslations, which are prevalent when $C(f(\hat{a}),f(a_e))$ dominates the sum as $P_\theta(\hat{a})$ is close to $0$ and $\beta$ is low or when $P_\theta(\hat{a})$ dominates the sum as $C(f(\hat{a}),f(a_e))$ is close to $0$ and $\beta$ is large. Further, the mean log probability term is unbounded, which makes finding a score threshold more challenging. Hence, we propose a novel scoring function $\mathcal{S}_g$ (Equation~\ref{eq:our_score}) that considers the squared geometric mean of $C(f(\hat{a}),f(a_e))$ and $P_\theta(\hat{a})$, to produce a bounded non-negative $(0,1)$ scoring function, which prioritizes both language generation and semantic translation objectives jointly, defined as:
\begin{equation} 
    \mathcal{S}_{g} = \underset{a_e}{\operatorname{argmax}}~[\underset{\hat{a}}{\operatorname{max}}~ \frac{C(f(\hat{a}),f(a_{e})) + 1}{2} \cdot e^{P_{\theta}(\hat{a})}]
    \label{eq:our_score}
\end{equation}
All re-prompting methods, by default, are reported using $\mathcal{S}_{w}$. We report results using $\mathcal{S}_{g}$ specifically for the re-prompting with explicit cause ($P_E$) method.

\subsection{Baseline: Plan Generation via Re-sampling}
\label{sec:resample}

 

When a plan action is not executable, the closed-loop re-sampling method does not use error feedback to generate corrective prompts. Instead the approach iteratively evaluates the top $k$ admissible actions proposed by the Planning LLM and grounded by the Translation LLM in reverse order of the weighted sum of mean log probability and cosine similarity until an executable action is found. If none of the $k$ re-sampled admissible actions are executable, plan generation terminates.
This ablation assesses whether CAPE's feedback allows for more efficient corrections due to the utility of precondition error information, rather than more attempts at proposing corrective actions.

\subsection{Baseline: Plan Generation with SayCan}
\label{sec:saycan}
We compare to SayCan~\cite{saycan_corl} as a baseline method. When generating every step, SayCan assigns a score for each action in the agent's repertoire and the action with the highest score is executed. This score is the product of the log probability and affordance for each action. 
This process is repeated until the termination skill (\code{done}) is assigned the highest score.
There are two important adjustments in our SayCan implementation for experiments in VirtualHome~\cite{puig2018virtualhome}:
\begin{itemize}
    \item As there are over $50$K possible object-action pairs in VirtualHome, it is intractable to evaluate every admissible skill for every step during planning. Instead the LLM generates a \textit{prototype} step. Using this we sub-sample the $500$ most semantically similar object-action pairs (measured by cosine similarity) and at most $1000$ object-action pairs containing the target object. This forms a subset of $\leq1500$ skills to iterate over and score. Subsampling semantically similar skills and matching to skills affecting the same objects ensures the $\leq$1500 subsampled skills also have the highest log probability according to the LLM. In most cases, nearly all the skills pertaining to a specific object are populated in the set of 1000, and additional semantically similar skills are added as part of the 500.
    \item A \textit{perfect} affordance model is initially used, since heuristic based precondition checks in VirtualHome allow $0\%$ affordance misclassification. However, as \citet{saycan_corl} mentions a $16\%$ of planning failure at minimum, where $35\%$ of these failures originate from errors related to the affordance model, we also present a \textit{noisy} ablation of SayCan with a $6\%$ ($16\%\times35\%$)  random chance of misclassifying the oracle affordance, i.e., false when actually true or true when actually false.
\end{itemize}
Similar to CAPE, SayCan assumes that language descriptions of an agent's skills are known and available during planning. SayCan leverages a trained affordance model (value function) to evaluate the executability of skills and can easily be extended to check for or predict language-specified precondition violations, similar to those leveraged in our method.


\section{Evaluation} 
We test the hypothesis that corrective re-prompting can increase the executability of LLM models for interpreting language directed to robots while maintaining plan correctness. We focus on larger state-of-the-art LLMs, particularly those in OpenAI's \code{davinci-instruct} line, for their demonstrated capabilities in instruction-following and planning tasks~\cite{brown2020language,summers-stay2021}. 
We evaluate eight approaches in a zero-shot setting: the three baselines -- \citet{huang2022language} (Section \ref{open-loop}), the closed-loop re-sampling (Section \ref{sec:resample}), and SayCan~\cite{saycan_corl} (Section \ref{sec:saycan}) -- and CAPE with our proposed ablations (Section~\ref{sec:reprompt}). We refer to CAPE's zero-shot approaches as success only ($\mathcal{Z}_{S}$), implicit cause ($\mathcal{Z}_{I}$), explicit cause ($\mathcal{Z}_{E}$), and explicit cause with scoring function ($\mathcal{Z}_{E} + {S}_g$). We also evaluate CAPE with explicit cause re-prompting in a few-shot setting ($\mathcal{F}_{E}$), with and without $S_g$, where we present the LLM with three examples of precondition errors and corresponding corrective actions to infer the appropriate corrective action for the target task.

\subsection{Experimental Setup}
We evaluate CAPE across seven scenes in VirtualHome~\cite{puig2018virtualhome}) and with a Boston Dynamics Spot robot (see Figure~\ref{fig:robot-demo}) using the metrics discussed in the following section. Our objective is to show that corrective re-prompting resolves unmet preconditions during planning and execution by embodied agents and robots in a variety of settings; VirtualHome provides a large skill sets with many objects, while the robot environments focus on physical embodiment with fewer objects and skills. For VirtualHome, we evaluate plans generated for $100$ household tasks (e.g., ``make breakfast", ``browse the Internet"). To show that our method can be extended to novel unstructured real-world environments, we compare plans generated by CAPE with those generated by the $3$ baselines across $6$ tasks for human-assistance and $2$ scenes for each task.




\begin{table*}[t]
    \centering
    \caption{Performance of baselines and CAPE across $100$ test-set task types and $7$ scenes in VirtualHome~\cite{puig2018virtualhome} (700 total).}
    \begin{tabular}{lcccccccc}
    \toprule[1pt]
    \textbf{Method} & \textbf{\%Correct}$\uparrow$ & \textbf{\%Exec.}$\uparrow$ & \textbf{\%Aff.}$\uparrow$ & \textbf{\%GS}$\uparrow$ & \textbf{LCS}$\uparrow$ & \textbf{Fleiss' Kappa}$\uparrow$ & \textbf{Steps}$\downarrow$ & \textbf{Corrections}$\downarrow$\\
    \toprule
    \textbf{Baselines}\\
    \citet{huang2022language} & 38.15 & 72.52 & 87.72 & 95.54 & 20.80 & 0.47 & 7.21 & N/A \\
    Re-sampling & 38.89 & 76.43 & 75.24 & 95.65 & 23.45 & 0.45 & 6.87 & 7.67 \\
    SayCan~\cite{saycan_corl} (Perfect) & 28.89 & \textbf{100.00} & \textbf{100.00} & 94.17  & 22.98 & 0.33 & 7.56 & N/A\\
    SayCan~\cite{saycan_corl} (Noisy) & 22.59 & 97.33 & 99.89 & 94.68 & 19.43 & 0.46 & \textbf{5.97} & N/A  \\
    \midrule
    \textbf{CAPE: Zero-Shot ($\mathcal{Z}$)} & & & \\
    Success Only ($\mathcal{Z}_{S}$) & 41.11 & 97.57 & 90.46 & 95.49 & 23.79 & 0.38 & 7.68 & 1.08 \\
    Implicit Cause ($\mathcal{Z}_{I}$) & 42.22 & 97.86 & 90.05 & 95.64 & 23.20 & \textbf{0.51} & 7.48 & 0.93 \\
    Explicit Cause ($\mathcal{Z}_{E}$) & 42.59 & 98.29 & 91.69 & 95.69 & 23.48 & 0.45 & 8.16 & \textbf{0.72} \\
    Explicit Cause ($\mathcal{Z}_{E}$ + $\mathcal{S}_g$) & 48.52 & \textbf{98.57} & 91.28 & 96.23 & 23.30 & 0.35 & 8.81 & 1.31 \\
    \midrule
    \textbf{CAPE: Few-Shot ($\mathcal{F}$)} & & & \\
    Explicit Cause ($\mathcal{F}_{E}$) & {47.04} & \textbf{98.57}  & \textbf{92.29} & 96.05 & \textbf{24.20} & 0.41 & 8.69 & 0.89 \\
    Explicit Cause ($\mathcal{F}_{E}$ + $\mathcal{S}_g$) & \textbf{49.63} & 96.29 &  90.93 & \textbf{96.29} & 23.47 & 0.39 & 9.35 & 1.82 \\
    \bottomrule[1pt]
    \end{tabular}
    \label{tab:results}
\end{table*}

\begin{table*}[t]
    \centering
    \caption{Performance of baselines and CAPE across $6$ test-set tasks and $2$ scenes for household tasks with robot demo (12 total).}
    \begin{tabular}{lccccccccc}
    \toprule[1pt]
    \textbf{Method} & \textbf{\%Correct}$\uparrow$ & \textbf{\%Exec.}$\uparrow$ & \textbf{\%Aff.}$\uparrow$ & \textbf{\%GS}$\uparrow$ & \textbf{LCS}$\uparrow$ & \textbf{Fleiss' Kappa}$\uparrow$ & \textbf{Steps}$\downarrow$ & \textbf{Corrections}$\downarrow$\\
    \toprule
    \textbf{Baselines}\\
    \citet{huang2022language} & 16.67 & 41.64 & 56.46 & 66.03 & 26.77 & 0.28 
    & \textbf{2.40} & N/A \\
    Re-sampling & 13.33 & 75.00 & 47.98 & 67.33 & 32.92 & \textbf{0.71} 
    & 4.60 & 13.19\\
    SayCan~\cite{saycan_corl} (Perfect) & 28.33 & 83.33 & \textbf{83.33} & 68.02 & 41.13 & 0.26 
    & 6.80 & N/A\\
    SayCan~\cite{saycan_corl} (Noisy) & 16.67 & 66.67 & 79.13  & 67.54 & 38.36 & 0.22 
    & 6.80 & N/A \\
    \midrule
    \textbf{CAPE: Zero-Shot ($\mathcal{Z}$)} & & & \\
    Success Only ($\mathcal{Z}_{S}$) & 18.33 & 75.00 & 43.05  & 66.02 & 32.45 & 0.28 
    & 3.04 & 2.25\\
    Implicit Cause ($\mathcal{Z}_{I}$) & 20.00 & 75.00 & 52.37 & 66.25 & 32.44 & 0.32 
    & 3.14 & 1.83\\
    Explicit Cause ($\mathcal{Z}_{E}$) & 31.67 & \textbf{100.00} & 79.69 & 69.18 & 48.12 & 0.11 
    & 6.30 & 1.91\\
    Explicit Cause ($\mathcal{Z}_{E} + \mathcal{S}_g$) & 23.33 & \textbf{100.00} & 79.04 & 69.85 & 46.68  & 0.12 
    & 6.30 & \textbf{1.73}\\
    \midrule
    \textbf{CAPE: Few-Shot ($\mathcal{F}$)} & & & \\
    Explicit Cause ($\mathcal{F}_{E}$) & 45.00 & \textbf{100.00} & 81.36 & \textbf{77.91} & 65.07 & 0.23 
    & 11.70 & 2.91\\
    Explicit Cause ($\mathcal{F}_{E} + \mathcal{S}_g$) & \textbf{50.00} & \textbf{100.00} & 80.70 & 77.40 & \textbf{69.77} & 0.12 
    & 11.30 & 2.90\\
    \bottomrule[1pt]
    \end{tabular}
    \vspace{-0.4cm}
    \label{tab:results-robot}
\end{table*}

\subsection{Robot Demonstration}
To demonstrate CAPE's capability on unstructured real-world tasks, we compare our re-prompting approaches against all $3$ baselines on the Boston Dynamics Spot, a quadruped robot with a single 6-DOF arm. The demonstrations use two novel scenes (a lab environment and a kitchen) with structural variation in the maps and objects in the environment. On average 9 household objects (e.g., phone, bed, coffee, etc.), each with five state attributes (e.g,
\code{location, grabbed, open, turned on}) are present in each scene. We evaluate performance on $6$ tasks: 1) Prepare for a run on a warm day, 2) Put the phone on the nightstand, 3) Iron a shirt, 4) Put mail in storage, 5) Organize Pantry, and 6) Put away groceries. We assume the Spot robot has access to a set of $14$ parametrized skills (e.g. \textit{stand up}, \textit{walk to}, \textit{pick up}, \textit{put}, \textit{touch}, \textit{look at}, \textit{open} and \textit{close}) and the initialization states (preconditions) needed for their execution. The robot first builds a semantic map from images taken and waypoints set across the scene; visual-language models (VLM) like (CLIP~\cite{clip} and CLIPSeg~\cite{luddeke2022clipseg}) are then used to ground admissible skills to spatial points for navigation or grasping in the physical environment, similar to approaches like NLMap-SayCan~\cite{chen2022nlmapsaycan}. The robot's embodiment (a single arm), a limited skill repertoire and extensibility to novel unstructured environments make this a challenging setting for task completion.  
%
Figure \ref{fig:robot-demo}  highlights how corrective prompting enables successful completion of the 
task \textit{"prepare for a run on a warm day"}.
Re-prompting enables the Spot to resolve precondition failures caused by the robot's initial state and due to its single-arm embodiment. We provide demonstrations for additional tasks and scenes in our supplementary video.

\subsection{Human Evaluation}
As in \citet{huang2022language}, we use human evaluation to determine the correctness of generated plans through the crowd-sourcing platform Prolific.\footnote{Prolific -- \url{https://www.prolific.co}} 50\% of the total tasks across all baselines and ablations were supplied to annotators. For each task, five annotators evaluate the grounded plan in English to determine whether it accomplishes the given task objective. Each plan is generated in a randomly selected environment.

\subsection{Evaluation Metrics}
\label{sec:metrics}



We adopt the \% Executability and \% Correctness metrics from \citet{huang2022language}.
\textbf{\% Executability} measures if \textit{all} grounded actions satisfy preconditions imposed by the environment i.e. if the \textit{entire} plan can be executed by the agent as afforded to its environment and state.
\textbf{\% Affordability} measures the average percentage of all plan steps that are executable, after skipping non-executable steps, in cases where the entire plan is not afforded execution (i.e. partial executability).

\textbf{\% Correct} is a human-annotated assessment of semantic correctness and relevance of a grounded plan to the target task. Assessing "quality" of natural language-based plans is difficult and potentially ambiguous using only executability i.e. an fully executable plan need not realize the task objective; thus, we conduct human evaluations where participants assign a binary score reflecting whether a plan is \textit{correct} or \textit{incorrect}. For a fairer representation of correctness, we account for executability constraints (i.e., precondition errors) by presenting human evaluators the plans up to the step where they remain executable by the agent for all methods (including baselines). 
Additionally, we report \textbf{Fleiss' Kappa} for \textit{\% Correct} inter-annotator agreement among participants in a categorical labeling task for our human annotations. This ranges from 0 to 1. Higher values indicate a stronger agreement between annotators~\cite{landis1977}.

\textbf{Longest Common Subsequence (LCS)} measures raw string overlap between generated grounded programs and the ground-truth programs as proposed by \citet{puig2018virtualhome}. LCS serves as a proxy for correctness as human evaluations more robustly measure plan semantics, i.e., human evaluations are not constrained by the richness of interactions in the embodied environment and variability of approaches to complete a task. 
We also report the average number of \textbf{Steps} and \textbf{Corrections} across tasks, which assess the total number of steps and corrective re-prompts/re-samples, respectively, needed to generate a plan. While these metrics are incidental to the goal (i.e. minimizing these metrics does not necessarily correlate to improved performance), they assess the relative efficiency of each prompting/sampling ablation towards correcting skill execution.
Finally, \textbf{Scene-Graph Similarity (\%GS)} reflects the percentage of state-object attributes that match between the final states resulting from execution of the generated grounded program ($\mathcal{G}_{gen}$) and the ground-truth human-written program ($\mathcal{G}_{gt}$). The number of matching attributes are normalized over the union of objects in both $\mathcal{G}_{gen}$ and $\mathcal{G}_{gt}$. This metric is invariant to differences in length and ordering of steps between generated and ground-truth plans, compared to a string-matching metric like LCS.

\section{Discussion}
\label{results}

\begin{figure}[h]
    \centering
    \includegraphics[width=1.0\columnwidth]{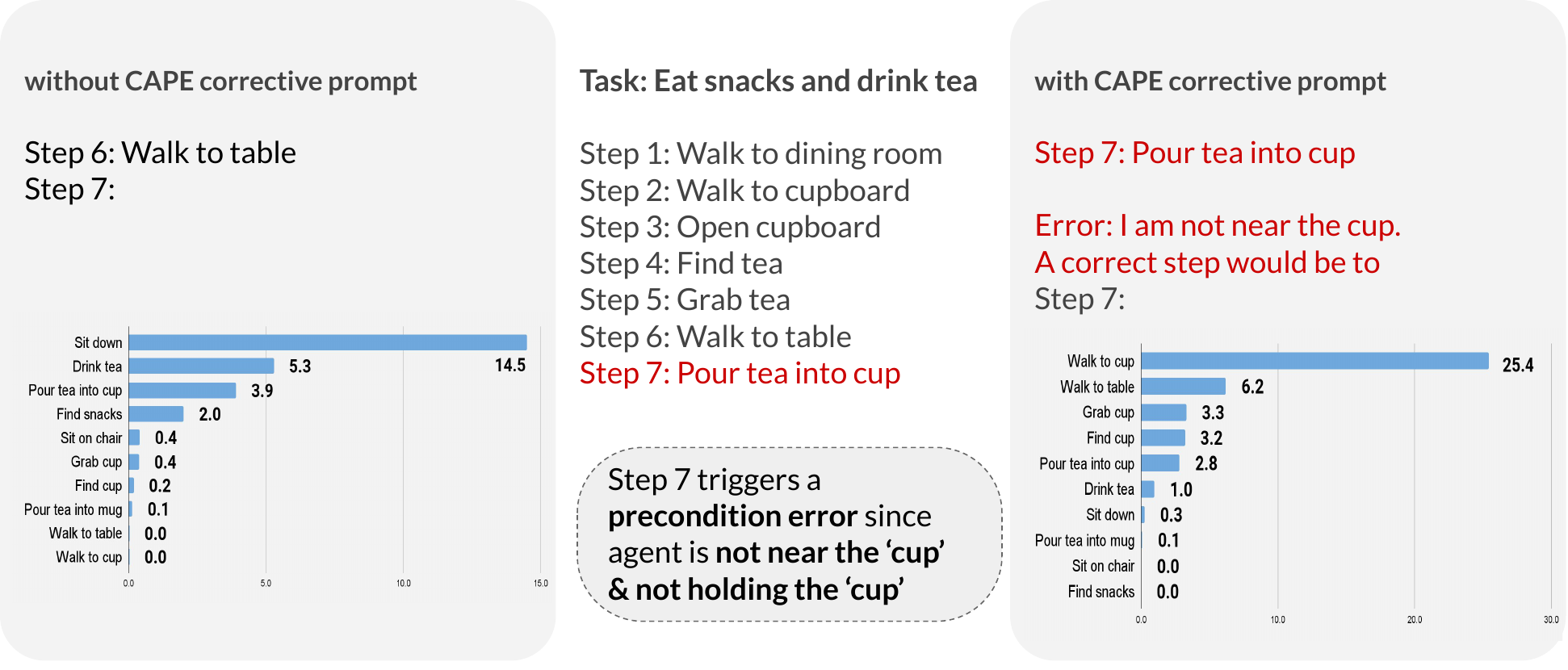}
    \caption{A qualitative example highlighting the impact of CAPE's corrective prompt on the Planning LLM's assigned probability distribution. CAPE's corrective prompt shifts the distributions towards actions that resolve preconditions and achieve the task objective}
    \vspace{-0.5cm}
    \label{fig:llm-distribution-shift}
\end{figure}

In VirtualHome~\cite{puig2018virtualhome}, CAPE generates plans that outperform competing methods (Table \ref{tab:results}). Our method \textbf{CAPE: Few-Shot with Explicit Cause} ($\mathcal{F}_{E}$ + $\mathcal{S}_g$) attains the highest combined performance for plan \textit{\% Correct} (49.63\%) and \textit{Executability} (96.29\%). For \% Correct, our method improves on SayCan (Perfect) by 71.80\% (absolute improvement of 20.74\%) while maintaining comparable executability and percentage of afforded steps, even though SayCan operates in an oracle setting with 0\% affordance misclassification. For all methods in Virtual Home experiments, the Fleiss' Kappa indicates moderate inter-annotator agreement for the \% Correct metric. Our zero-shot ablations with varying specificity of error information outperform the SayCan and \citet{huang2022language} baselines as well as other baselines (\citet{progprompt}) that report 90\% executability and 72\% graph similarity on 77 Virtual Home tasks using 3 in-context examples with \code{davinci-codex} model. This demonstrates the effectiveness of our method even without few-shot learning. The results also show that increasing the specificity of error information improves the performance of CAPE.
Our method's plans are also higher quality while requiring fewer \textit{Corrections} than the Re-Sampling baseline, which indicates the added utility of corrective actions from precondition error information. Our method also outperforms SayCan across nearly all metrics, even though SayCan implicitly assumes additional environment feedback in the form of a trained affordance model. Furthermore, our method significantly reduces time complexity over SayCan, $O(n)$ compared with $O(|s|^{n})$ respectively, where $s$ is the skill repertoire and $n$ the number of plan steps, since SayCan iterates the entire skill space before generating every step.

We present the results of the robot demonstration in Table-\ref{tab:results-robot}. Our method \textbf{CAPE: Few-Shot with Explicit Cause} ($\mathcal{F}_{E}$ + $\mathcal{S}_g$) attains the highest \textit{Executability} (100\%) due to re-prompting with precondition errors. Our method also shows improvement in combined performance for plan \textit{\% Correct} (50\%) 
and \textit{Executability} (96.29\%). For \% Correct, our method improves upon SayCan (Perfect) by 76.49\% (absolute improvement of 21.67\%) while attaining comparable percentage of afforded steps, even though SayCan operates in an oracle setting and is guaranteed to produce executable skills. SayCan usually fails because the affordance function "funnels" (severely limits) the available actions, sometimes leading plans into local optima i.e. afforded actions with highest log-probability do not resolve precondition errors that are critical to task completion and afforded actions that do resolve these precondition do not have sufficient log-probability.
For all methods, the Fleiss' Kappa indicates modest inter-annotator agreement between annotators for the \% Correct metric, except for Re-Sampling where annotators unanimously agree that the generated plans do not successfully complete the task.

\begin{figure}[th]
    \centering
    \includegraphics[width=1.0\columnwidth]{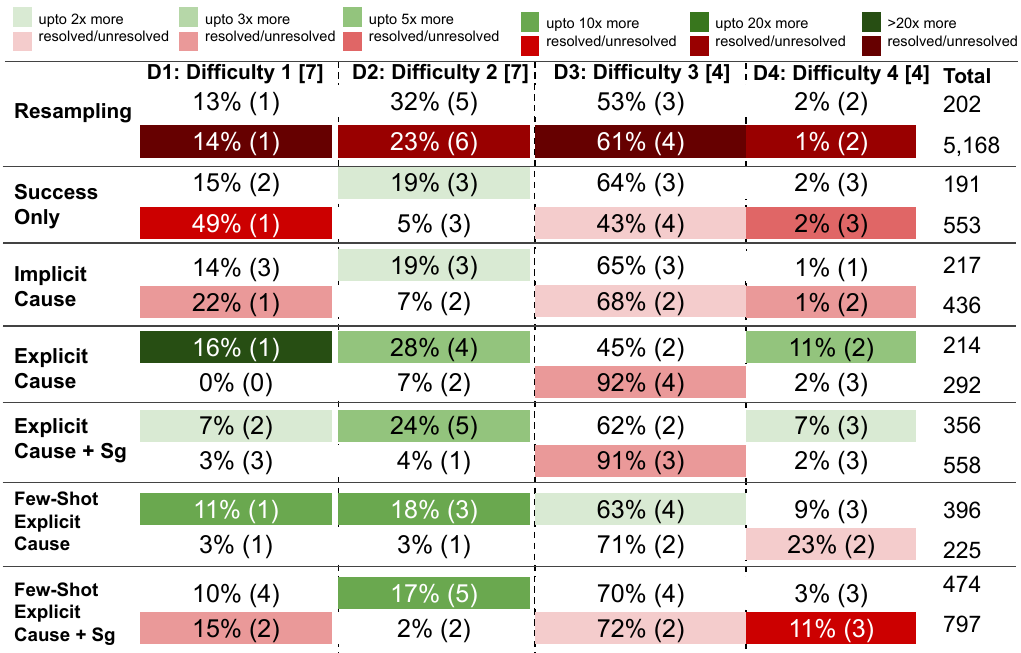}
    \caption{The distribution of precondition errors that are resolved (top) and unresolved (bottom) for all reprompting methods, across 4 difficulties. Values in bracket show the no. of error types for a given difficulty}
    \vspace{-0.5cm}
    \label{fig:error-analysis}
\end{figure}

Finally, more explicit CAPE ablations resolve a larger proportion of more difficult precondition errors. Figure \ref{fig:error-analysis}
shows the distribution of $22$ VirtualHome precondition error types across $4$ difficulty levels for all CAPE and resampling ablations. Difficulty levels include errors that require: no corrections (D1: e.g., ``opening an open door''), one-step corrections (D2), multi-step corrections (D3), and long-term planning with ambiguous resolution (D4: e.g., ``too many objects on the table''). More difficult errors require broader historical/environment context to resolve. 
A precondition error in step $i$ is `resolved' only if the plan progresses to step $\geq i+1$ before the next error. There are $5$ observations: (1) majority of resolved/unresolved errors in all ablations fall under D3; (2) $\mathcal{F}_{E}$ is the only ablation with more resolved (396) than unresolved (225) errors and an average of $4\times$ more resolutions across difficulties; (3) re-sampling has $25\times$ more unresolved errors with a minimum of $20\times$ more non-resolutions across difficulties; (4) increased error specificity can more readily resolve D1--3 errors, with sharpest increase for D2 errors; (5) whilst $\mathcal{S}_{g}$ disproportionately increases total number of unresolved errors, diluting the ratio of resolved errors, $\mathcal{S}_{g}$ also maintains the \textit{proportion} of unresolved errors in each difficulty and broadens the diversity of resolved errors compared to unresolved errors. 

As shown in Figure \ref{fig:llm-distribution-shift}, CAPE's corrective prompts shift the Planning LLM's assigned probability distribution towards natural-language tokens (that ground to actions) that resolve precondition errors. Not only is the assigned probability distribution shifted, but higher relative weightage is assigned to actions that resolve precondition errors compared to those that do not -- due to CAPE's corrective prompts.

\section{Related Work}

\label{relatedwork}
\textbf{Large Language Models for Task Planning:}
%
Works that are significantly related to our paper are \citet{huang2022language}, SayCan~\cite{saycan_corl}, and \citet{gramopadhye2022generating}, which integrate LLMs into an open-loop planning pipeline. \citet{huang2022language} use a prompting strategy to derive step-by-step plans that achieve the goal presented in a prompt. Our work extends their approach by incorporating feedback from the environment as an auxiliary input to improve the executability of derived plans.\citet{gramopadhye2022generating}  also improves upon the \citet{huang2022language} by providing environmental context to the LLM to generate contextually suitable plans.

\citet{saycan_corl} introduces SayCan, a LLM-integrated pipeline that proposes a sequence of actions to achieve specific goals grounded to affordance with a predefined set of robot-executable skills (all demonstrated by an expert) using semantic similarity from language prompt.
However, these works only implicitly incorporate "feedback" by selecting actions that are visually afforded in the current state. They do not address action failure or failure recovery.

\textbf{Visual \& Language Feedback for Planning:} Following our prior work~\cite{raman2022planning}, recent works have shown the efficacy of LLM-based autonomous agents in leveraging language feedback for reasoning about errors~\cite{guo2023doremi,zhang2023grounding,shinn2023reflexion,liu2023reflect}. Reflexion~\cite{shinn2023reflexion} converts scalar feedback (from heuristic-based evaluators) into structured linguistic feedback with long-term memory to improve decision making via trial-and-error; in contrast, CAPE does not enable multiple trials nor access retrospective feedback to re-plan from initialization. CAPE only utilizes the agent's current action history and does not assume access to long-term feedback over multiple episodes. Other works such as DoReMi~\cite{guo2023doremi}, \citet{zhang2023grounding} also assume access to a set of primitive skills but combine VLMs and LLMs to detect action failures by monitoring properties associated with constraints (either from planning domains or proposed by LLM) for the skills being executed. DoReMi \cite{guo2023doremi} focuses on low-level failure recovery and assumes the LLM has direct access to additional information (e.g., the entire skill repertoire, skills' constraints, task instructions) whilst CAPE provides implicit feedback to the LLM for specific skill preconditions. \citet{zhang2023grounding} also use VLMs to verify action affordances based on preconditions extracted from PDDL and track updated environment state after skill execution, which is provided to the LLM during next step generation. Environment state information is stored external to the agent in CAPE: the LLM used by CAPE does not directly have access to the underlying state and only receives implicit feedback in the form of re-prompts with which the LLM has to infer the current state and propose an appropriate next step. Additionally, both methods assume VLMs have access to the global visual state during skill execution in order to detect failures, which may not translate naturally to the environments and embodiment types we study, i.e., simulated and real-world agents that have partial observability and use egocentric image feedback. 
REFLECT \cite{liu2023reflect} utilizes multi-modal feedback to extract a hierarchy of events and visually informed scene graphs, which are then used to explain failures during planning. However, assessing object states from visual and auditory feedback requires pre-defining audio labels and object state labels for visual/audio grounding, also requiring a non-trivial amount of extra effort in addition to pre-defining all skills.

\textbf{Task and Motion Planning:}
In task and motion planning (TAMP), robot planning and execution processes are decoupled in a hierarchical manner~\cite{kaelbling2010hierarchical,garrett2021integrated}.
This involves the integration of \textit{task planning}, which aims to find a sequence of actions that realize state transitions and goal state corresponding to a high-level problem~\cite{ghallab2016automated}, and \textit{motion planning}, which aims to find physically consistent and collision-free trajectories that realize the objectives of a task plan~\cite{lozano1979algorithm,dornhege2009integrating}.
Instead of relying on explicitly defined structures or symbols as typically used in TAMP, LLMs can provide an agent or robot with an implicit representation of action and language, allowing it to interpret a task and identify key details (such as objects or actions) that are related to the problem at hand.



\textbf{Commonsense Knowledge in LLMs:}
Other works explore the degree to which LLMs contain commonsense world knowledge. The Winograd Schema Challenge \cite{levesque2012winograd} and WinoGrande benchmark \cite{sakaguchi2021winogrande} evaluate commonsense reasoning in word problems. The Winoground dataset~\cite{thrushwinoground2022} investigates commonsense reasoning in a related image caption disambiguation challenge. LLMs have improved upon baseline methods for this task \cite{brown2020language} indicating that language model scale contributes to commonsense reasoning performance. Our system supports the finding that LLMs contain latent commonsense world knowledge sufficient to improve plan executability given precondition errors.

\section{Conclusion}
\label{conclusion}
\label{futurework}

We propose CAPE, a re-prompting strategy for LLM-based planners, which injects contextual information in the form of precondition errors, parsed from environment feedback, which substantially improves the executability and correctness of LLM-generated plans and enables agents to resolve action failure. Our experiments in VirtualHome \cite{puig2018virtualhome} and on the robot demonstration show that corrective prompting results in more semantically correct plans with fewer precondition errors than those generated by baseline LLM-planning frameworks (\citet{huang2022language} and SayCan~\cite{saycan_corl}) and re-sampling. CAPE overcomes the computational intractability of applying SayCan to environments with large numbers of agent skills. CAPE enables more executable and correct plans in less time, while exploring a narrower subset of the skills and using far fewer interjections.

\subsection{Limitations}
\label{limitations}

CAPE achieves strong competitive performance over baseline methods by leveraging a minimal but efficient architecture while only receiving implicit uni-modal (linguistic) feedback from the environment. However, we acknowledge several limitations of CAPE:

\textbf{Relaxing precondition assumption}: CAPE can be more flexible by restricting the assumption that precondition propositions with language feedback are known. Incorporating methods to automatically ground preconditions to binary questions (like \citet{zhang2023grounding}) could allow CAPE to automatically detect or predict the cause of skill failures using additional prompts; furthermore, utilizing LLMs to generate preconditions for future actions (e.g., deriving grounded constraints using methods like the constraint generation module in DoReMi~\cite{guo2023doremi}) could allow CAPE to scale efficiently to larger action spaces and define parametrized dependencies or constraints for skills that are not manually defined.


\textbf{Open-Query Error Handling}: Methods like REFLECT~\cite{liu2023reflect} have shown that grounding feedback from multiple modalities enables LLMs to reason about causes of skill failure. This approach leverages a multi-modal which could allow CAPE to verify action affordances and generate prompts in an open-query style for a wider range of error types than the ones specified by the precondition definition. Multi-modal feedback can even be used upon successful skill execution to allow CAPE to update an internal structured representation of the current environment state, which can be used to determine the affordance of future actions without having to encode all environment state transitions. 

\textbf{Correcting Perception \& Low-level Control}: To control the influence of low-level skill (perception, joint manipulation, end effector) errors, CAPE abstracts low-level control into a repertoire of high-level skills that we assume execute perfectly for the purpose of high-level planning. The same abstraction is applied to baselines as well, such that only logical pre-condition errors (the focus of our work) can disrupt plan execution. Several works (SayCan \cite{saycan_corl}, NLMap-SayCan \cite{chen2022nlmapsaycan}, \citet{huang2022language} and Inner Monologue \cite{huang2022inner}) make similar assumptions on high-level skills, though integrating failure detection and recovery for low-level control (DoReMi \cite{guo2023doremi}) could enable CAPE to more robustly recover from additional failure-types.


\section*{Acknowledgements}
This work is supported by ONR under grant award numbers N00014-21-1-2584 and N00014-22-1-2592, NSF under award number CNS-2038897, and with support from Echo Labs. 
Additionally, this material is based upon work supported by the Defense Advanced Research Projects Agency (DARPA) under Contract No. HR001122C0007. Any opinions, findings and conclusions or recommendations expressed in this material are those of the author(s) and do not necessarily reflect the views of the Defense Advanced Research Projects Agency (DARPA).


\bibliographystyle{IEEEtranN}
\bibliography{references}  

\appendix
\section{\textbf{Appendix}}
\label{appendix}

\subsection{Error Types in VirtualHome}

To assess the viability of engineered corrective prompts, we qualitatively analyze and categorize the $22$ precondition error types in Virtual Home across $4$ categories of increasing difficulty. Since `corrective prompts' are to generate corrective actions that resolve preconditions, it becomes vital to assess the nature of precondition errors and how they might be corrected for. The errors types and their difficulty classifications are listed in Tables ~\ref{tab:all_error_types} ~\ref{tab:all_error_types_2} and ~\ref{tab:all_error_types_3} below. 

\label{error-types}
\begin{table*}[h!]
    \fontsize{8}{9}\selectfont
    \centering
    \begin{tabular}{lp{2.2cm}p{5cm}p{4cm}}
        \toprule[1pt] 
        \textbf{Precondition Error} & \textbf{Difficulty Level} & \textbf{Description} \\
        \toprule[1pt] 
        X is not movable & 1 & The \code{move} action does not apply to object \code{X}. Difficulty 1 since unmovable objects are governed by environment dynamics (not a planning failure) so the plan can proceed without needing to resolve any preconditions\\
        \midrule
        X cannot be opened & 1 & The \code{open} action does not apply to object \code{X}. Difficulty 1 since unopenable objects are governed by environment dynamics (not a planning failure) so the plan can proceed without needing to resolve any preconditions\\   
        \midrule
        X is not cuttable & 1 & The \code{cut} action does not apply to object \code{X}. Difficulty 1 since cuttable objects are governed by environment dynamics (not a planning failure) so the plan can proceed without needing to resolve any preconditions\\ 
        \midrule
        X is not a receptacle & 1 & The \code{put down} action cannot be done onto object \code{X} since it is not a receptacle. Difficulty 1 since receptacle attribute is governed by environment dynamics (not a planning failure) so the plan can proceed without needing to resolve any preconditions\\
        \midrule
        X is not lookable & 1 & The \code{look} action does not apply to object \code{X}. Difficulty 1 since lookable attribute is governed by environment dynamics (not a planning failure) so the plan can proceed without needing to resolve any preconditions\\
        \midrule
        agent is already sitting & 1 & The \code{sit} action cannot be performed since the agent is already sitting down. Difficulty 1 because the preconditions for the action are already satisfied and no further preconditions need to be resolved\\
        \midrule
        X is sitting & 1 & The failed action has a precondition requiring the agent to be standing, which is not satisfied. Difficulty 1 because the \code{stand up} only needs to be executed, which itself has no preconditions i.e. \code{stand up} is universally afforded execution\\
        \midrule
        X is not lying or sitting & 2 & The failed action has a precondition requiring the agent to be sitting or lying down, which is not satisfied. Difficulty 2 because either the \code{sit} or \code{lie down} actions need to be executed, which themselves at least $1$ precondition e.g. \code{sit} requires the agent to be near a sittable object and not already sitting/lying down\\
        \midrule
        X is turned off / closed twice & 2 & The \code{turn on} or \code{open} action cannot be performed since object \code{X} is already turned on (or open). Difficulty 2 because preconditions for the action are already satisfied but the `corrective action' (\code{turn off} or \code{close}) is not universally executable\\
        \midrule
        X is turned on / opened twice & 2 & The \code{turn off} or \code{close} action cannot be performed since object \code{X} is already turned off (or closed). Difficulty 2 because preconditions for the action are already satisfied but the `corrective action' (\code{turn on} or \code{open}) is not universally executable\\
        \midrule
    \end{tabular}
    \caption{Description of error types observed in the VirtualHome environment.}
    \label{tab:all_error_types}
\end{table*}

\label{error-types}
\begin{table*}[h!]
    \fontsize{8}{9}\selectfont
    \centering
    \begin{tabular}{lp{2.2cm}p{5cm}p{4cm}}
        \toprule[1pt] 
        \textbf{Precondition Error} & \textbf{Difficulty Level} & \textbf{Description} \\
        \toprule[1pt] 
         X is not closed & 2 & The failed action has a precondition requiring object \code{X} to be closed, which is not satisfied. Difficulty 2 because a single action \code{close X} needs to be executed to resolve the precondition error, which has at least $1$ precondition e.g. agent needs to be near object \code{X} object \code{X} must be open\\
         \midrule
         X is not open (or not openable) & 2 & The failed action has a precondition requiring object \code{X} to be open, which is not satisfied. Difficulty 2 because a single action \code{open X} needs to be executed to resolve the precondition error, which has at least $1$ precondition e.g. agent needs to be near object \code{X} and object \code{X} must be closed\\
        \midrule       
 agent is not facing X & 2 & The failed action has a precondition requiring agent to turn towards object \code{X}, which is not satisfied. Difficulty 2 because a single action \code{turn to X} needs to be executed to resolve the precondition error, which has at least $1$ preconditions e.g. agent needs to be near object \code{X}\\
        \midrule
        X is not grabbed  & 3 & The failed action has a precondition requiring agent to be holding object \code{X}, which is not satisfied. Difficulty 3 because a multiple actions need to potentially be executed to resolve the precondition error (\code{walk to X}, \code{put down Y} - object being held currently, \code{grab X}) i.e. the agent needs to be near object \code{X} and have at least $1$ free hand\\
        \midrule
        agent is not holding X  & 3 & The failed action has a precondition requiring agent to be holding object \code{X}, which is not satisfied. Difficulty 3 because a multiple actions need to potentially be executed to resolve the precondition error (\code{walk to X}, \code{put down Y} - object being held currently, \code{grab X}) i.e. the agent needs to be near object \code{X} and have at least $1$ free hand\\
        \midrule
        agent is not holding anything & 3 & The failed action has a precondition requiring agent to be holding a (specific) object, but the agent is not holding anything. Difficulty 3 because precondition resolution requires multiple corrective actions e.g. inferring what object to grab, \code{walk to} object, \code{grab} object\\
        \midrule
        agent is not close to X & 3 & The failed action has a precondition requiring the agent to be near object \code{X}, which is not satisfied. Difficulty 3 because precondition resolution requires multiple corrective actions e.g. \code{walk to X}, \code{open door} in case door in path is closed\\

        \midrule
        X is inside another closed object & 4 & The failed action has a precondition requiring object \code{X} to be accessible (removed) from within another closed object, which is not satisfied. Difficulty 4 because precondition resolution requires long-term planning with corrective actions that themselves have at least $1$ precondition e.g. inferring the closed object, \code{open Y} the object that is closed, \code{grab X} to remove from within object \code{Y}\\

\end{tabular}
    \caption{Description of error types observed in the VirtualHome environment.}
    \label{tab:all_error_types_2}
\end{table*}

\label{error-types}
\begin{table*}[h!]
    \fontsize{8}{9}\selectfont
    \centering
    \begin{tabular}{lp{2.2cm}p{5cm}p{4cm}}
        \toprule[1pt] 
        \textbf{Precondition Error} & \textbf{Difficulty Level} & \textbf{Description} \\

\midrule
        too many things on X & 4 & The failed action has a precondition requiring the receptacle \code{X} to be empty or below maximum capacity, which is not satisfied. Difficulty 4 because precondition resolution requires contextualization and long-term planning with corrective actions that themselves have at least $1$ precondition e.g. inferring objects consuming capacity on object \code{X}, grabbing said objects, walking to another empty receptacle, placing said objects on receptacles\\

        \midrule
        agent does not have a free hand & 4 & The failed action has a precondition requiring $1$ of  the agent's hand to be free i.e. not holding or grabbing an object, which is not satisfied. Difficulty 4 because precondition resolution requires contextualization and long-term planning with corrective actions that themselves have at least $1$ precondition e.g. inferring the objects in the agent's hands, walking to another empty receptacle, placing objects on empty receptacles.\\

        \midrule
        door between room \code{X} and room {Y} is closed & 4 & The failed action has a precondition requiring the door connecting rooms \code{X} and \code{Y} to be open, which is not satisfied. Difficulty 4 because precondition resolution requires contextualization with corrective actions that themselves have at least $1$ precondition e.g.  inferring which door instance is closed, walking to door, opening door\\

\end{tabular}
    \caption{Description of error types observed in the VirtualHome environment.}
    \label{tab:all_error_types_3}
\end{table*}

\subsection{Causes for execution errors with open-loop planning}
\label{more-errors}

\begin{figure*}[t]
    \centering    
    \includegraphics[scale=0.45]{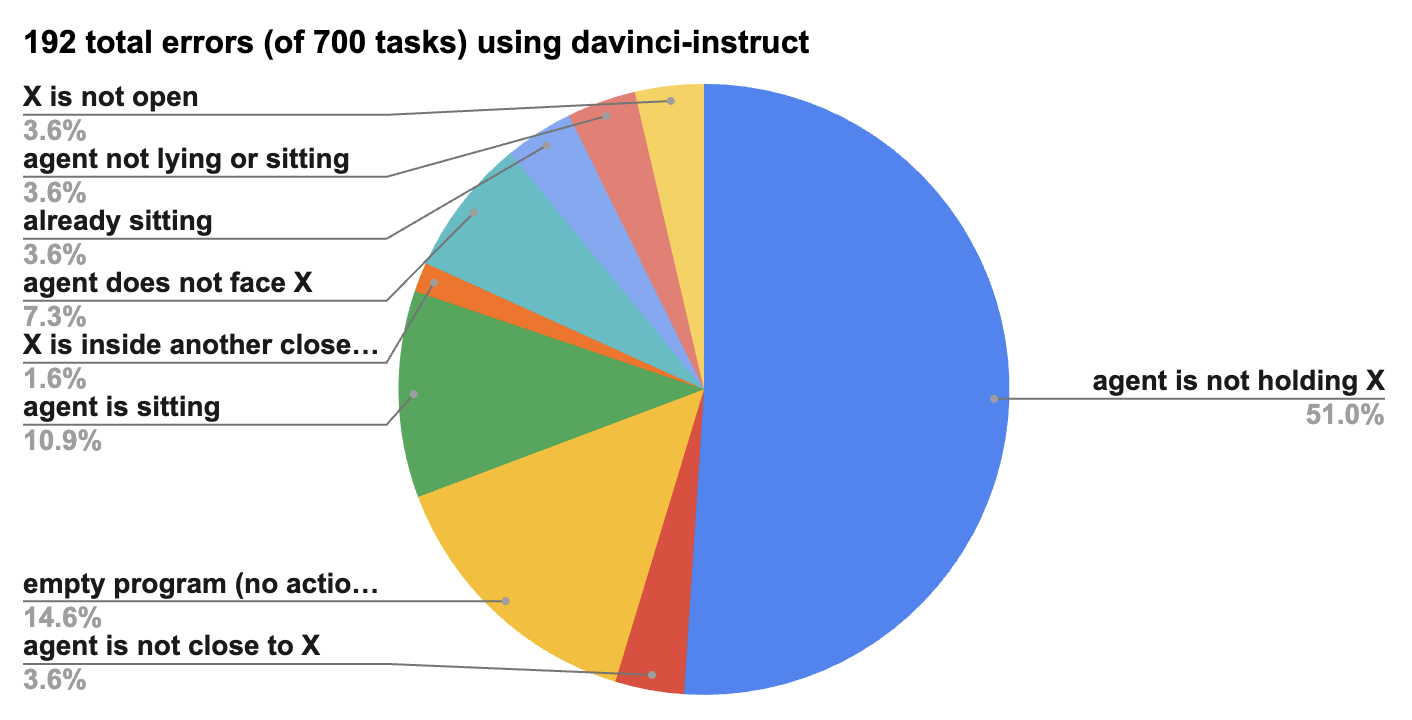}
    \caption{Decomposition of precondition errors into error types when running \citet{huang2022language} baseline on \code{davinci-instruct} LLM}
    \label{fig:error-composition}
\end{figure*}

Figure~\ref{fig:error-composition} highlights the percentage composition of the $192$ errors observed when running the \citet{huang2022language} baseline (on 700 task scene combinations) using the \code{davinci-instruct} LLM line.

The ``agent is not holding X'' error is the most prominent precondition error type, accounting for a majority ($51\%$) of errors. This is followed by the  ``empty program" ($14.6\%$) and ``agent is sitting'' ($10.9\%$) precondition errors. The significant presence of ``empty program'' errors indicates that a rescoring of generated actions (using scoring function $\mathcal{S_{g}}$ with the Planning LLM) may be needed in addition to corrective prompting. Of the $9$ error types observed, the majority lie between difficulty $2$ and $3$, indicating that correcting most precondition errors for the \citet{huang2022language} baseline requires just $1$ or up to a few ($2-3$) corrective actions.

Our qualitative observations find that certain tasks are \textit{intrinsically} more complex, increasing the difficulty of precondition errors and the number of preconditions to satisfy via corrective prompting. For example, the baseline failed on the task ``Keep cats inside while door is open" because the \code{door} was closed in the initial state. 
There are some qualitative observations that validate this:
\begin{itemize}
    \item Tasks such as ``Entertain" are generally vague, which could impede the LM's ability to generate sensible actions without accessing feedback about the environment's current state.
    \item Certain tasks enforce implications on the environment's initial state (e.g,. tasks like ``push all chairs in``  require no actions or precondition resolution since all chairs begin tucked under a table). Therefore, viable plans requires feedback about the initial environment state.
    \item The Planning LLM generating steps without environment feedback is usually very optimistic, inferring object locations from the task/prompt and assuming objects are readily available. Therefore,  without environment feedback, generated plans are  not aligned with the constraints of the embodied environment.
\end{itemize}  

\subsection{Hyper-parameter search for LLMs in closed-loop domain}
\label{param-search}
Our approaches (Sections \ref{sec:reprompt} and \ref{sec:resample}) form a closed-loop system by providing a Planning LLMs with precondition error information (using environment feedback) to generate corrective actions. It is clear that including additional re-prompts and corrective actions would increase plan length and the frequency of tokens used, however the optimal LLM hyper-parameters for the closed-loop domain are unclear. Thus we perform a hyper-parameter sweep to optimize the Planning LLM's parameter selection, using the re-sampling baseline (see Section \ref{sec:resample}) as a proxy for all closed-loop approaches i.e. re-sampling, SayCan and CAPE re-prompting.

Our hyper-parameter sweep is performed across different temperatures and presence penalties in the ranges shown in Table~\ref{tab:hyper-params} below. These parameters, respectively, influence how out of distribution a proposed action might be and penalize the repetition of previously generated actions (topics), making them strong levers for Planning LLM performance in the closed-loop domain.

\begin{table}[h]
\centering
\caption{The range of hyper-parameter (temperature and presence penalty) values explored as part of parameter sweep}
\renewcommand{\arraystretch}{1.1}
\begin{tabular}{cc}
    \toprule
    \textbf{Hyper-parameter} & \textbf{Search Values}\\ 
    \midrule
    \textit{Temperature} & 0.3, 0.5, 0.7 \\  \textit{Presence Penalty} & 0.3, 0.5, 0.7 \\ 
    \bottomrule
\end{tabular}
\label{tab:hyper-params}
\end{table}

The columns in Table~\ref{tab:hyper-params} represent presence penalties (PresPen) and the rows represent temperature (Temp). Each table element evaluates a set of metrics from top to bottom for the particular temperature-presence penalty combination: executability, \% executed, graph similarity, LCS-EP, number of steps and the number of corrections (as from Section \ref{sec:metrics}).

\begin{table*}[h]
    \centering
    \renewcommand{\arraystretch}{1.25}
    \begin{tabular}{ c|ccc } 
    \toprule
    & PresPen@0.3 & PresPen@0.5 & PresPen@0.7 \\
    \midrule
    \multirow{6}{*}{Temp@0.3} 
    & 75.03\% & 74.43\% & 74.15\% \\ 
    & 73.81\% & 73.24\% & 72.96\%  \\ 
    & 22.35 & 22.42 & 22.44\\ 
    & 6.37 & 6.33 & 6.37\\ 
    & 7.60 & 7.59 & 7.62 \\
    \midrule
    \multirow{6}{*}{Temp@0.5} 
    & 75.72\% & 76.43\% & \textbf{77.15\%} \\ 
    & 74.44\% & 75.24\% & \textbf{75.86\%}\\
    & 23.28 & \textbf{23.45} & 23.38 \\ 
    & 6.90 & 6.87 & 6.93 \\ 
    & 7.87 & 7.67 & 7.79\\
    \midrule
    \multirow{6}{*}{Temp@0.7} 
    & 75.23\% & 75.29\% & 75.15\% \\ 
    & 74.04\% & 74.20\% & 73.99\% \\ 
    & 23.53 & 22.69 & 22.68\\
    & 7.23 & 7.21 & 7.18 \\ 
    & 8.831 & 8.78 & 8.66 \\ 
    \bottomrule
    \end{tabular}
    \caption{Hyper-parameter search performance (Executability, \% Afforded, LCS, no. steps and no. corrections) over different temperatures (Temp) and presence penalties (PresPen) using the \code{instruct-davinci} LLM.}
    \label{tab:instruct-davinci-hyperparams}
\end{table*}

\textbf{Presence Penalty} does not seem to influence executability at higher temperatures; for lower temperatures, executability decreases with increased presence penalties. The number of corrective steps and total steps appear to be independent of presence penalty as well.

\textbf{Temperature} and executability seem to have an `optimal temperature' that maximizes executability, around which (for higher or lower temperatue) executability is lower, for all presence penalties tested. A similar correlation is seen between LCS and temperature as well as affordance and temperature with peak LCS/executability at a temperature value of $0.5$. Additionally, the number of corrective steps and total steps seem to both monotonically increase with temperature.

Overall, as per Table~\ref{tab:instruct-davinci-hyperparams}, the maximum executability and \% affordance is observed using presence penalty and temperature of $0.7$ and $0.5$. Maximum LCS, however, is observed using presence penalty $0.5$ and temperature of $0.5$. Since affordance filtering (SayCan baseline) and corrective prompts would naturally induce higher executability and affordance, we deem LCS a more challenging metric for improved performance. Thus we use $0.5$ presence penalty and temperate for our zero-shot re-prompting, few-shot re-prompting, re-sampling baseline and SayCan baseline.

\subsection{Qualitative Examples of Generated Plans}

We present the generated plans for $10$ sample tasks comparing the \citet{huang2022language} baseline with the re-sampling baseline, CAPE's zero-shot and few-shot ablations. 

We find qualitative improvements in the planner's corrective capabilities (in resolving precondition errors) as well as improved specificity and alignment with the task's objective.

\begin{table*}[h]
    \small
    \centering
    \renewcommand{\arraystretch}{1.25}
    \begin{tabular}{ p{2cm}p{4.3cm}p{4.3cm}p{4.3cm} } 
    \toprule
    \textbf{Task} & \textbf{Baseline Method}\newline \citep{huang2022language} & \textbf{Re-Sampling} & \textbf{Re-Prompting} \\
    \midrule
    Get glass of milk & 
    
    Step 1: Walk to dining room\newline
    Step 2: Walk to kitchen counter\newline
    Step 3: Find kitchen counter\newline
    Step 4: Open stove\newline
    Step 5: Find milk\newline
    Step 6: Grab milk\newline
    Step 7: Close stove\newline
    Step 8: Walk to fridge\newline
    Step 9: Find fridge\newline
    Step 10: Open fridge & 

    Step 1: Walk to dining room\newline 
    Step 2: Walk to kitchen counter\newline 
    Step 3: Find kitchen counter\newline 
    Step 4: Find fridge\newline 
    Step 5: Open fridge\newline 
    Step 6: Find milk\newline 
    Step 7: Grab milk & 

    Step 1: Walk to dining room \newline
    Step 2: Walk to kitchen counter\newline
    Step 3: Find kitchen counter\newline
    Step 4: Find fridge\newline
    Step 5: Open fridge\newline
    Step 6: Find milk\newline
    Step 7: Grab milk\\ 

    \midrule

    Prepare pot of boiling water & 
    
    Step 1: Find stove\newline
    Step 2: Find pot\newline
    Step 3: Turn to stove & 
    
    Step 1: Find stove\newline 
    Step 2: Find pot\newline
    Step 3: Turn to stove\newline  & 
    
    Step 1: Find stove\newline
    Step 2: Find pot\newline
    Step 3: Turn to stove\newline
    Step 4: Switch on stove\newline
    Step 5: Put pot on stove\newline
    Error: I don't have the pot. A correct step would be to\newline
    Step 5: Find pot\\
    \midrule

    Look at mirror & 
    
    Step 1: Walk to bedroom\newline
    Step 2: Walk to mirror\newline
    Step 3: Find mirror\newline
    Step 4: Turn to mirror\newline
    Step 5: Look at mirror & 
    
    Step 1: Walk to bedroom\newline
    Step 2: Walk to mirror\newline
    Step 3: Find mirror\newline
    Step 4: Turn to mirror\newline 
    Step 5: Look at mirror & 
    
    Step 1: Walk to bedroom\newline
    Step 2: Walk to mirror\newline
    Step 3: Find mirror\newline
    Step 4: Turn to mirror\newline
    Step 5: Look at mirror\newline
    Step 6: Find face\newline
    Step 7: Turn to face\newline
    Step 8: Look at face\newline
    Step 9: Find hair\newline
    Step 10: Turn to hair\newline
    Step 11: Look at hair\newline
    Step 12: Find clothes dress\newline
    Step 13: Turn to clothes dress\newline
    Step 14: Look at clothes dress \\

    \midrule

    Listen to Music & 

    Step 1: Walk to home office\newline
    Step 2: Walk to remote control\newline
    Step 3: Find remote control\newline
    Step 4: Grab remote control\newline
    Step 5: Walk to couch\newline
    Step 6: Sit on couch\newline
    \textit{Error: agent is sitting so cannot execute Step 8 until the agent stands up}\newline
    Step 7: Touch remote control\newline
    Step 8: Find television\newline
    Step 9: Switch on television\newline
    Step 10: Turn to television\newline
    Step 11: Watch television\newline
    Step 12: Switch on stereo\newline
    Step 13: Turn to stereo\newline
    Step 14: Switch on cd player\newline
    Step 15: Turn to cd player & 
    
    Step 1: Walk to home office\newline
    Step 2: Walk to remote control\newline
    Step 3: Find remote control\newline
    Step 4: Grab remote control\newline
    Step 5: Walk to couch\newline
    Step 6: Sit on couch\newline
    Step 7: Touch remote control\newline
    Step 8: Find music stand &
    
    Step 1: Walk to home office\newline
    Step 2: Walk to remote control\newline
    Step 3: Find remote control \newline
    Step 4: Grab remote control \newline
    Step 5: Walk to couch\newline
    Step 6: Sit on couch\newline
    Step 7: Touch remote control\newline
    Step 8: Find television\newline
    Error: I am not near the television. A correct step would be to\newline
    Step 8: Walk to tv\newline
    Error: I am sitting. A correct step would be to\newline
    Step 8: Stand up\newline
    Step 9: Walk to stereo\newline
    Step 10: Switch on stereo\newline
    Step 11: Turn to stereo\newline
    Step 12: Find cd\newline
    Step 13: Grab cd\\

\bottomrule
\end{tabular}
\caption{Qualitative examples of generated plans for three cases: 1) the baseline method~\cite{huang2022language}, 2) our re-sampling method, and 3) CAPE's reprompting ablations}
\end{table*}

\begin{table*}[h]
    \small
    \centering
    \renewcommand{\arraystretch}{1.25}
    \begin{tabular}{ p{2cm}p{4.3cm}p{4.3cm}p{4.3cm} } 
    \toprule
    \textbf{Task} & \textbf{Baseline Method}\newline \citet{huang2022language} & \textbf{Re-Sampling} & \textbf{Re-Prompting} \\
    \midrule
        Hang pictures &

    Step 1: Walk to home office\newline
    Step 2: Walk to table\newline
    Step 3: Find keys\newline
    Step 4: Grab keys\newline
    Step 5: Walk to wall\newline
    Step 6: Find hanger\newline
    Step 7: Put keys on hanger\newline
    Step 8: Walk to livingroom\newline
    Step 9: Find picture\newline
    \textit{Error: agent has not picked up picture, so cannot execute Step 10}\newline
    Step 10: Put picture on hanger &

Step 1: Walk to livingroom\newline 
Step 2: Walk to table\newline 
Step 3: Find picture\newline
Step 4: Grab picture\newline
Step 5: Walk to wall\newline
Step 6: Find hanger\newline
Step 7: Put picture on hanger &
Step 1: Walk to livingroom\newline
Step 2: Walk to table\newline
Step 3: Find picture\newline
Step 4: Grab picture\newline
Step 5: Walk to wall\newline
Step 6: Find hanger\newline
Step 7: Put picture on hanger\\
\midrule
Eat snacks and drink tea  &
Step 1: Walk to livingroom\newline
Step 2: Sit on couch\newline
Step 3: Drink tea\newline
Step 4: Eat food snack\newline
Step 5: Drink juice
&
Step 1: Walk to livingroom\newline
Step 2: Walk to couch\newline
Step 3: Walk to kitchen\newline
Step 4: Walk to cupboard\newline
Step 5: Open cupboard\newline
Step 6: Find tea\newline
Step 7: Grab tea\newline
Step 8: Walk to kettle\newline
Step 9: Switch on kettle\newline 
&
Step 1: Walk to dining room\newline
Step 2: Walk to cupboard\newline
Step 3: Open cupboard\newline
Step 4: Find tea\newline
Step 5: Grab tea\newline
Step 6: Walk to table\newline
Step 7: Pour tea into cup\newline
Error: I am not near the cup. A correct step would be to\newline
Step 7: Walk to cup\newline
Step 8: Find cup\newline
Step 9: Grab cup\newline
Step 10: Pour tea into cup\newline
Step 11: Drink tea\newline
...\\
\midrule
Receive credit card  & Step 1: Walk to home office & Step 1: Walk to home office\newline
Step 2: Find chair\newline
Step 3: Sit on chair & Step 1: Walk to home office\newline
Step 2: Walk to cabinet\newline
Step 3: Find cabinet\newline
Step 4: Open cabinet\newline
Step 5: Find creditcard\newline
Step 6: Grab creditcard\\
   \bottomrule
    \end{tabular}
    \caption{Qualitative examples of generated plans for three cases: 1) the baseline method~\cite{huang2022language}, 2) our re-sampling method, and 3) CAPE's reprompting ablations
    }
\end{table*}

\begin{table*}[h]
    \small
    \centering
    \renewcommand{\arraystretch}{1.25}
    \begin{tabular}{ p{2cm}p{4.3cm}p{4.3cm}p{4.3cm} } 
    \toprule
    \textbf{Task} & \textbf{Baseline Method}\newline \citep{huang2022language} & \textbf{Re-Sampling} & \textbf{Re-Prompting} \\
    \midrule
Empty dishwasher and fill dishwasher & Step 1: Walk to dishwasher\newline
Step 2: Open dishwasher & Step 1: Walk to dishwasher\newline
Step 2: Open dishwasher\newline
Step 3: Walk to sink\newline
Step 4: Find faucet\newline
Step 5: Switch on faucet\newline
Step 6: Find bowl\newline
Step 7: Grab bowl\newline
Step 8: Rinse bowl
& Step 1: Walk to dishwasher\newline
Step 2: Open dishwasher\newline
Step 3: Walk to sink\newline
Step 4: Find faucet\newline
Step 5: Switch on faucet\newline
Step 6: Find bowl\newline
Step 7: Grab bowl\newline
Step 8: Rinse bowl\newline
Step 9: Switch off faucet\newline
Error: I am not near the faucet. A correct step would be to\newline
Step 9: Walk to faucet\\
\midrule
Take shoes off & 
Step 1: Walk to home office\newline
Step 2: Walk to shoes\newline
Step 3: Find shoes & 
Step 1: Walk to home office\newline 
Step 2: Walk to shoes\newline 
Step 3: Find shoes & 
Step 1: Walk to home office\newline
Step 2: Walk to shoes\newline
Step 3: Find shoes\newline
Step 4: Take off shoes\newline
Step 5: Walk to dresser\newline
Step 6: Open dresser\newline
Step 7: Grab shoes\newline
Step 8: Put shoes in dresser\newline
Error: putin dresser is not allowed. A correct step would be to\newline
Step 8: Put shoes in closet\\
    \bottomrule
    \end{tabular}
\caption{Qualitative examples of generated plans for three cases: 1) the ~\cite{huang2022language} baseline, 2) our re-sampling baseline, and 3) CAPE's reprompting ablation.
}
\end{table*}

\end{document}